\documentclass{llncs}  

\usepackage{graphicx} 
\usepackage[T1]{fontenc}
\usepackage{amsmath}
\usepackage[english]{babel}
\usepackage{array,tabularx}
\usepackage{booktabs} 
\usepackage{makeidx}  
\usepackage{xparse}
\usepackage{moredefs}
\usepackage{lips}
\usepackage{epstopdf} 
\usepackage{color}
\usepackage{csquotes}
\usepackage{xcolor}
\usepackage{times}
\usepackage[hidelinks]{hyperref}
\usepackage[nolist,nohyperlinks]{acronym}
\usepackage{comment}
\usepackage{paralist}
\usepackage{cleveref}
\usepackage{wrapfig}
\usepackage{multirow}
\usepackage{microtype}
\usepackage{listings}
\usepackage[super]{nth}
\usepackage{natbib}

\usepackage{float}
\floatstyle{plaintop}
\restylefloat{table}

\usepackage[style=base,labelfont=bf,labelsep=period,tableposition=top,font=small]{caption}
\makeatletter
\renewcommand\@biblabel[1]{#1.}
\makeatother
\addto\captionsenglish{}

\usepackage{fancyhdr}
\fancypagestyle{WI_footer}{\fancyhf{}\fancyfoot[L]{\color{black}\scriptsize 19th International Conference on Wirtschaftsinformatik\\September 2024, Würzburg, Germany}}

\crefformat{footnote}{#2\footnotemark[#1]#3}
\expandafter\def\expandafter\UrlBreaks\expandafter{\UrlBreaks
  \do\a\do\b\do\c\do\d\do\e\do\f\do\g\do\h\do\i\do\j%
  \do\k\do\l\do\m\do\n\do\o\do\p\do\q\do\r\do\s\do\t%
  \do\u\do\v\do\w\do\x\do\y\do\z\do\A\do\B\do\C\do\D%
  \do\E\do\F\do\G\do\H\do\I\do\J\do\K\do\L\do\M\do\N%
  \do\O\do\P\do\Q\do\R\do\S\do\T\do\U\do\V\do\W\do\X%
  \do\Y\do\Z}

\newcolumntype{L}[1]{>{\raggedright\arraybackslash}p{#1}}   
\newcolumntype{C}[1]{>{\centering\arraybackslash}p{#1}}     
\newcolumntype{R}[1]{>{\raggedleft\arraybackslash}p{#1}}    

\begin{document}
\frontmatter          

\mainmatter              

\title{Getting in Contract with Large Language Models – An Agency Theory Perspective on Large Language Model Alignment}

\subtitle{Research Paper} 
\author{Sascha Kaltenpoth\inst{1} \and Oliver Müller\inst{1}}

\institute{Paderborn University, Department of Business Administration and Economics, Paderborn, Germany \\
\email{\{sascha.kaltenpoth,oliver.mueller\}@uni-paderborn.de}}

\maketitle
\setcounter{footnote}{0}

\begin{abstract}
Adopting Large language models (LLMs) in organizations potentially revolutionizes our lives and work. However, they can generate off-topic, discriminating, or harmful content. This AI alignment problem often stems from misspecifications during the LLM adoption, unnoticed by the principal due to the LLM's black-box nature. While various research disciplines investigated AI alignment, they neither address the information asymmetries between organizational adopters and black-box LLM agents nor consider organizational AI adoption processes. Therefore, we propose \textit{LLM ATLAS} (LLM \textbf{A}gency \textbf{T}heory-\textbf{L}ed \textbf{A}lignment \textbf{S}trategy) a conceptual framework grounded in agency (contract) theory, to mitigate alignment problems during organizational LLM adoption. We conduct a conceptual literature analysis using the \textbf{organizational LLM adoption phases} and the \textbf{agency theory} as concepts. Our approach results in (1) providing an extended literature analysis process specific to AI alignment methods during organizational LLM adoption and (2) providing a first LLM alignment problem-solution space. \\

\end{abstract}

\thispagestyle{WI_footer}

\section{Introduction}
Adopting highly capable Large language models (LLMs) such as Llama 3 \citep{Meta2024} or GPT-4 \citep{openai2023gpt4} in organizational processes, e.g., as domain-specific conversational agents, has the potential to revolutionize our lives and work \citep{Feuerriegel2023}. However, they can generate off-topic, discriminating, or harmful content under certain circumstances \citep{Teubner2023, touvron2023llama}. In this case, the artificial intelligence (AI) agent's actual behavior differs from that intended by its principal, e.g., the user or designer. This \textit{AI alignment problem} \citep{Gabriel2020, hadfield2019} often stems from misspecifications during the LLM adoption (e.g., implementation of LLM-driven systems into the organizational infrastructure) \citep{shen2023large, wang2023aligning}, unnoticed by the principal due to the LLM's black-box nature.

This information asymmetry between the adopting organization and the LLM agent possibly results in serious consequences, such as lawyers facing sanctions for citing fake cases \citep{Bohannon2023} or failures in chatbot services \citep{Moench2024}, for which users usually blame the company rather than the LLM-based system itself \citep{Pavone2022}. Therefore, it is crucial for organizations to mitigate information asymmetries during LLM adoption and ensure that these models are aligned with human, organizational, and social values \citep{askell2021general, kenton2021alignment}. 

Various research disciplines investigated AI alignment, particularly LLM alignment. Machine learning (ML) researchers focused on technical solutions, such as high-quality demonstration training data \citep{Ouyang2022, zhou2023lima}, incorporating external knowledge \citep{Lewis2020, zhang2024raft}, output verification \citep{Li2023, openai2023gpt4}, or alignment benchmarks \citep{Zha2023, chang2024}. AI ethics researchers analyzed possible harms and provide governance recommendations \citep{Henderson2018, weidinger2021ethical, Mkander2023}. Furthermore, multiple literature reviews explained, categorized, and assessed LLM alignment approaches \citep{ji2024ai, shen2023large, wang2023aligning}.

While these studies provide some guidance for organizations that strive to adopt LLMs, they neither address the information asymmetries between the organizational adopter and the black-box LLM agent nor do they explicitly consider organizational AI adoption processes. Like most AI-based approaches, LLMs' organizational adoption requires selection, training, evaluation, deployment, and monitoring of the model \citep{Kreuzberger2023, Symeonidis2022}.

Therefore, we propose \textit{LLM ATLAS} (LLM \textbf{A}gency \textbf{T}heory-\textbf{L}ed \textbf{A}lignment \textbf{S}trategy) a conceptual framework grounded in agency (contract) theory, to mitigate alignment problems during organizational LLM adoption. Highlighting the information asymmetry between the adopting organization and the black-box LLM, we adapt the idea of \citet{hadfield2019} to frame LLM alignment as an agency problem. We conduct a conceptual literature analysis \citep{Schryen2015} using the combination of the \textbf{organizational LLM adoption phases} and the \textbf{agency theory} to analyze the literature. While we draw on existing AI and LLM alignment literature reviews instead of search and assessment, we synthesize a problem-solution mapping for LLM alignment during the organizational LLM adoption by applying agency theory to all organizational LLM adoption phases.

LLM ATLAS provides a simple approach to identifying agency problems in the organizational LLM adoption process and synthesizing solutions from the literature. Our approach extends the existing AI alignment literature and supports organizational LLM adoption by (1) providing an extended literature analysis process specific to AI alignment methods during organizational LLM adoption and (2) providing a first proof-of-concept of the problem-solution space derivation. In future research, we plan to extend this initial literature analysis with a comprehensive literature review and an extendable format for the presentation of the problem-solution space in the form of a multi-contributor website. We hope to encourage other researchers to participate in the research on the intersection of LLM alignment and organizational LLM adoption.

\section{Background}
\subsection{Large Language Models - A black-box Architecture}
Large language models (LLMs) such as GPT-4 or LLaMA, are neural networks, usually based on a decoder-only Transformer structure \citep{openai2023gpt4, touvron2023llama}, that autoregressively predict the next token $t_{n+1}$ based on the conditional probability distribution $P(t_{n+1}|t_1...t_n)$, where $t_1...t_n$ refers to the context sequence of tokens \citep{Shanahan2023}. They contain up to billions of parameters and are (pre-)trained on vast amounts of textual data \citep{bommasani2022opportunities, Shanahan2023}. These characteristics enable them to be pre-trained to solve many cognitive tasks, such as answering questions, generating coherent, meaningful texts, and passing various human exams at a human-like level \citep{Teubner2023, chang2024}. Furthermore, pre-trained LLMs possess foundational knowledge and can be adapted to various downstream tasks and domains \citep{bommasani2022opportunities}. Therefore, they are often called foundation models. However, foundation models are black-box architectures in terms of their neural network architecture and their vast amount of training data \citep{bommasani2022opportunities}. Consequently, the learned representations and their alignment to human, organizational, and social values are not visible to a potential adopter \citep{askell2021general}, causing an information asymmetry between the LLM and its adopter.

\subsection{The AI and LLM Alignment Problem}
The AI alignment problem refers to situations where an AI’s observed behavior differs from that intended by its designers or expected by its users \citep{Gabriel2020, hadfield2019}. This problem typically originates from information asymmetries between the AI (agent) and its designers and users (principal) \citep{hadfield2019}. A main alignment concern in the context of LLMs is hallucinations, that is, factually incorrect or unfaithful responses \citep{ji2023surveyhallucination}. Another common concern is social, gender-, or ethical biases of LLMs, especially foundation models \citep{weidinger2021ethical, kenton2021alignment}. LLM alignment aims to align LLMs with the task, on the one hand, and human, organizational, and social values, on the other hand, \citep{askell2021general, kenton2021alignment}. This broad goal can be operationalized drawing on the definition of \citet{askell2021general}, stating that an appropriately aligned LLM agent needs to commit to helpfulness, honesty, and harmlessness. While helpfulness refers to an agent being able to solve human-given tasks, honesty refers to giving accurate and correct information and signaling confidence in the given answer or solution \citep{askell2021general}. Harmlessness refers to not being offensive, discriminatory, biased, and refusing dangerous behavior.

\subsection{Existing Frameworks} \label{relatedwork}
The research on LLMs is studied in two comprehensive literature reviews \citep{minaee2024large, zhao2023survey}. \cite{zhao2023survey} provide an overview of the development of LLMs within the past decade. The authors differentiate between closed- and open-source LLMs, investigating their data sources and training strategies. In contrast, \cite{minaee2024large} categorize LLMs by their capabilities. They divide LLMs into basic, pre-trained LLMs, emerging, further fine-tuned LLMs, and augmented LLMs, which can use tools and APIs \citep{minaee2024large}. Both reviews consider alignment as a technique to train LLMs. While \cite{minaee2024large} provide an overview of hallucination evaluations and consider alignment as a common LLM training strategy, \cite{zhao2023survey} provide an overview of possible alignment datasets, alignment tuning strategies, and evaluation criteria, such as the helpfulness, honesty, harmlessness by \cite{askell2021general}.

\cite{ji2024ai} comprehensively review AI alignment problems and approaches. They identify four essential criteria for AI alignment: robustness, interpretability, controllability, and ethicality (RICE). Robustness refers to the resilience of AI systems, especially the correctness of outputs. In contrast, interpretability refers to the honesty of LLMs and using interpretability tools to investigate LLM reasoning \citep{ji2024ai}. While controllability refers to human supervision and intervention ability, if the model deviates from the expected behavior, ethicality states that the system should adhere to society’s norms and values \citep{ji2024ai}. \cite{Memarian2023} investigate the fairness, accountability, transparency, and
ethics (FATE) framework on AI in the higher education literature. Their findings show a tendency to use FATE in descriptive rather than technical terms and also that FATE is investigated in quantitative rather than qualitative means \citep{Memarian2023}.

\cite{shen2023large} and \cite{wang2023aligning} both provide a comprehensive survey of LLM alignment. While \cite{wang2023aligning} focus on human value-driven alignment, including human-generated data, human preferences in training, and human evaluation strategies, \cite{shen2023large} provides an overview of human- and non-human-driven alignment approaches categorized by inner alignment, outer alignment, and interpretability. Outer alignment refers to choosing the right loss or reward functions and ensuring the match of training objective and human intention. In contrast, inner alignment ensures that the LLMs' inner representations match the goals after training \citep{shen2023large}. The authors refer to interpretability as methods, models, and tools that facilitate humans to understand the LLM's inner workings \citep{shen2023large}.

Besides the comprehensive literature reviews on LLMs, AI, and LLM alignment, \cite{hadfield2019} refer to the agency theory and incomplete contracting as viewpoints to analyze the problems of AI alignment, especially reward misspecification based on incomplete information. They state that reward misspecification is a fundamental problem of AI alignment and suggest incorporating human decisions into the AI alignment.

Considering all related studies, technical solutions address various AI and LLM alignment problems. However, for an organization to adopt and appropriately align LLMs, it is crucial to understand the organizational LLM adoption phases and the asymmetries between an LLM and its adopter to decide on the proper alignment strategy to consider for every step of the adoption process.

\section{Methodology}
We apply an initial conceptual literature analysis \citep{Schryen2015}. Based on the stated similarities of LLM alignment and agency theory, we combine agency theory with the organizational LLM adoption phases as concepts to analyze the LLM alignment problems and solutions during organizational LLM adoption. In the following, the \textbf{organizational LLM adoption phases}, the \textbf{agency theory}, and \textbf{their connection to the LLM alignment} are motivated as categorization concepts for the literature analysis, followed by a description of our conceptual literature analysis procedure.

\subsection{Categorization Concepts}
\subsubsection{Alignment Problems in Organizational LLM Adoption}
We derive the organizational LLM adoption processes from three common ML and AI adoption processes: The CRISP-DM (cross-industry process for data mining) \citep{martinez2021crisp20, wirth2000crisp}, MLOps (machine learning operations) \citep{Kreuzberger2023, Symeonidis2022}, and LLMOps (large language model operations) \citep{Kulkarni2023}. All frameworks broadly comprise a (1) business problem definition, (2) data acquisition and preparation, (3) model selection, (4) model development, comprising training and evaluation, and (5) deployment and monitoring, which are organized mainly by data scientists, ML engineers, or other IT-related decision-makers \citep{aryan2023costly, Kreuzberger2023, Symeonidis2022}. The following refers to the IT-related decision-maker as a data scientist.

\textit{Business problem definition} usually starts with requirements gathering and engineering \citep{Kreuzberger2023, wirth2000crisp}. Usually, the data scientist gathers the requirements from several stakeholders and determines the downstream task to solve. As LLMs usually enable solving multiple downstream tasks, several requirements determine the considerable LLM type, such as foundation models, conversational, instruction-following, or tool-LLMs \citep{Kulkarni2023}. 

\textit{Data acquisition and preparation} is highly dependent on the business problem to solve \citep{testi2022, Symeonidis2022}. Based on the requirements, the data scientist needs to acquire data. Many datasets exist to train LLMs \citep{minaee2024large}. However, most organizations prefer to generate problem-specific training data based on organizational documents or human demonstrations \citep{Gururangan2020, liu2024chipnemo, Ouyang2022}, possibly combining it with publicly available domain-specific data \citep{HuggingFace2020, wu2023bloomberggpt}. Whether the data scientist acquires problem-specific data or utilizes publicly available datasets, ensuring proper alignment of the datasets is crucial to prevent model bias \citep{wang2023aligning} possibly leading to racial and sexist responses such as a python function that determines a good scientist as male and white \citep{Teubner2023}.

\textit{Model selection} comprises the selection of appropriate model candidates, training them on the acquired public and problem-specific data \citep{wirth2000crisp, testi2022}. The selection is possible from various open-source LLMs on platforms like HuggingFace Hub \citep{HuggingFace2020}. However, closed-source (commercial) LLMs such as GPT-3, GPT-4, or PALM2 \citep{anil2023palm, brown2020gpt3, openai2023gpt4} are also possible candidate models. The data scientist must ensure that the pre-trained or fine-tuned LLM is appropriately aligned to avoid accidentally incorporating biased or hallucinating models into organizational applications \citep{bommasani2022opportunities}.

\textit{Model development} of LLMs comprises the model training and evaluation. Pre-trained LLMs (i.e., foundation models) are adopted by the organization \citep{Kulkarni2023}. Thus, the data scientist must align the LLM to the objectives intended by the organization. Prompting is a common paradigm to adopt and align LLMs \citep{liu2023prompt}. It describes the process of passing downstream tasks as prompts, well-crafted, unambiguously defined instructions in human language, to LLMs to generate responses without further LLM training \citep{liu2023prompt}. In contrast, prompt engineering denotes the prompt development process, which also supports domain adaptation and fine-tuning LLMs \citep{Gururangan2020, liu2023prompt, zhou2023lima}.

During domain-adaptive pre-training (DAPT), an LLM is further trained on domain-specific textual data to infuse domain knowledge into the model \citep{Gururangan2020}. Another strategy to incorporate domain knowledge is retrieval-augmented generation (RAG) \citep{Lewis2020}. During RAG, documents similar to the prompt are retrieved first, followed by prompting the model to respond based on the documents \citep{Lewis2020}. Whereas RAG and DAPT mainly use raw domain-specific data such as product reviews or internal documents, supervised fine-tuning (SFT) leverages high-quality demonstration data from human labelers \citep{Ouyang2022, zhou2023lima}. This demonstration data is used to fine-tune the LLM, improving the (domain-specific) LLM capabilities and alignment \citep{openai2023gpt4, touvron2023llama2}. Furthermore, human preference incorporation further enhances the model outputs \citep{Christiano2017, Ouyang2022}, but will be outlined in Section \ref{sec:results}.

Evaluation of LLMs is possible using diverse benchmark datasets \citep{chang2024}. While overarching benchmarks evaluate capabilities in natural language understanding and generation, such as the HELM (Holistic Evaluation of Language Models) benchmark \citep{liang2023holistic}, various domain-specific benchmarks exist \citep{chang2024}. Recently, a shift from overarching benchmarks to more challenging task-specific benchmarks is observable \citep{chang2024}. Still, an organizational LLM adoption usually requires evaluating the problem or task within the specific domain \citep{liu2024chipnemo, wu2023bloomberggpt}. Thus, a data scientist must determine the right evaluation benchmarks and strategies, especially to ensure an appropriate alignment \citep{chang2024}.

\textit{Model deployment and monitoring} depends on the LLM user. LLMs can be deployed using external APIs, platforms, or self-hosted integrated solutions such as ERP-system integrations \citep{HuggingFace2020, Gallardo2023gpt3api, Kulkarni2023}. Users possibly are the organization's customers in conversational service agents \citep{Feuerriegel2023} but also employees using internal organizational assistant system \citep{liu2024chipnemo}. The data scientist must consider the LLM's end user to decide on the deployment strategy, such as API, web interface, or self-hosted integrated solution \citep{Gallardo2023gpt3api, Kulkarni2023}. Additionally, the LLM responses must be monitored to prevent a lack of up-to-date information \citep{openai2023gpt4} and prevent insecure outputs \citep{openai2023gpt4, touvron2023llama2}.

\subsubsection{Agency Theory as LLM Alignment Lense}
Agency theory describes problems of information asymmetry \citep{Akerlof1970} similarly to AI alignment \citep{hadfield2019}. An agency problem occurs in every relationship where a principal mandates an agent, and the principal cannot gain all information about the agent's characteristics and actions \citep{Eisenhardt1989}, possibly leading to a divergence in their preferred actions based on the utility maximization of both parties \citep{Jensen1976}. The agent's behavior can differ from the behavior intended by its principal. Information asymmetry can arise ex-ante as hidden characteristics or ex-post as hidden actions \citep{Linder2013}.

\textit{Hidden characteristics} about the agent are not initially known by the principal \citep{Akerlof1970, Spence1973}. \citet{Akerlof1970} gives the example of a used car market, where the buyer (principal) cannot determine the quality of a car offered by a seller (agent), leading to adverse selection with high high-quality car sellers leaving the market as they gain no buyer. Similarly, an organization (principal) cannot determine a pre-trained LLM's (agent's) alignment degree without further information due to the black-box architecture \citep{bommasani2022opportunities}.

\textit{Screening and signaling} aim to reduce the information asymmetry between the principal and the agent \citep{Spence1973, Wolpin1977}. In screening, the principal spends additional effort to gain information about the agent, e.g., an organization can derive information about a pre-trained LLM before adoption \citep{Mitchell2019}. Signaling refers to the opposite strategy, where the agent provides additional information to the principal. For example, assuming that additional education results in enhanced job productivity, an appropriate mechanism for screening and signaling can be an educational certification in the form of degrees \citep{Akerlof1970, Wolpin1977}. At the same time, an LLM benchmark can serve as a screening mechanism for LLMs \citep{chang2024}.

\textit{Hidden actions} occurs when the principal cannot verify the agent's actions after contracting \citep{Linder2013}, possibly leading to moral hazard if the agent differs from the expected behavior to maximize its own utility \citep{Eisenhardt1989, hokstoem1979}, e.g., project managers who work mainly without supervision and must meet schedules. The project managers would probably rather deliver low-quality software than delay the schedule \citep{Tuttle1997}. An unproperly aligned LLM-based conversational agent possibly answers to an insecure prompt, e.g., a request for a guideline to build a bomb, instead of refusing an answer \citep{openai2023gpt4}.

One solution can be \textit{monitoring} the agent to unravel its actions, like management boards or LLM output verification. Contractual \textit{bonding} in the form of \textit{incentives} indirectly prevents opportunistic behavior by the agent \citep{hokstoem1979}. An example of bonding by incentives is the contractual bonus. If a bonus for project managers is based on user satisfaction, it ``bonds'' the project manager to deliver high-quality software instead of delivering low-quality software fast to meet the schedule \citep{Tuttle1997}.

\subsection{Literature Analysis}
\begin{figure}
  \includegraphics[width=\linewidth]{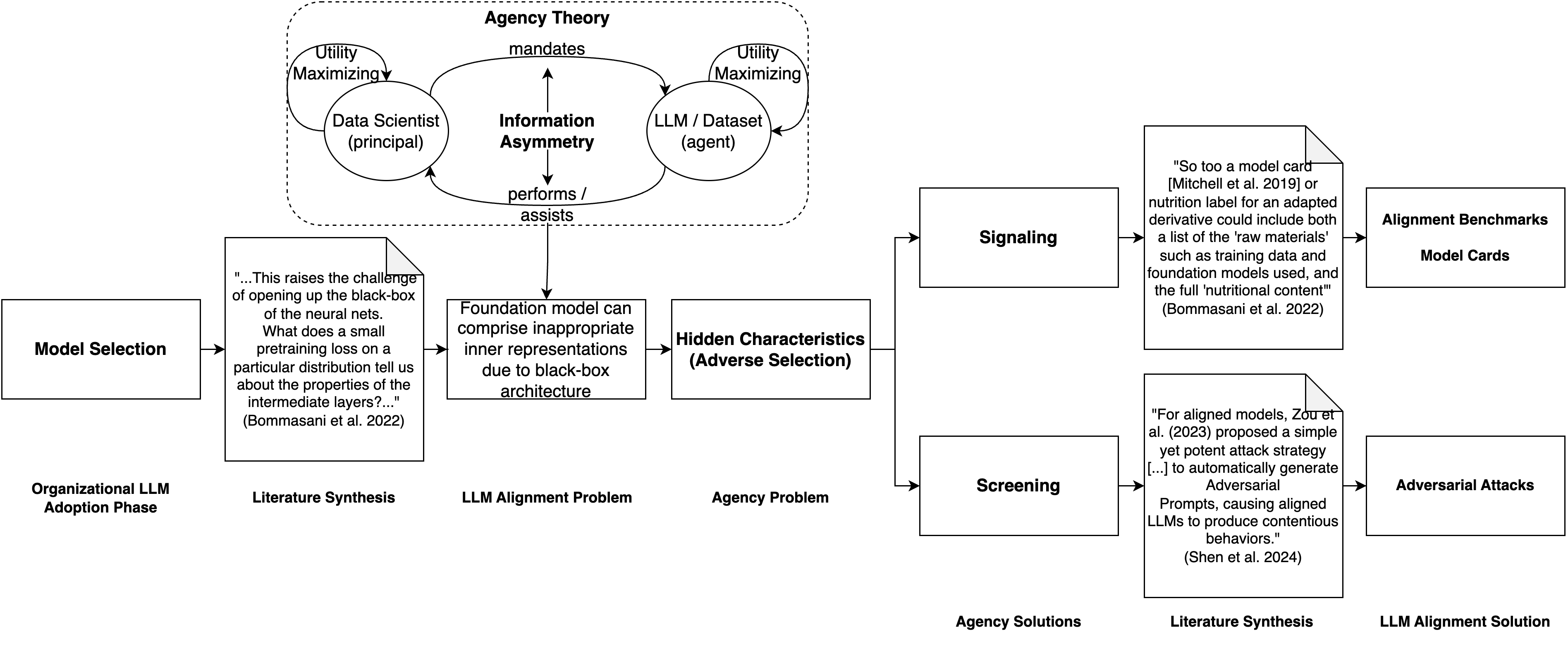} 
  \caption{The LLM ATLAS Methodology} 
  \label{fig:methodology}
\end{figure}

For our initial literature analysis, we adjust the common literature analysis process \citep{Schryen2015, Webster2002} and draw on already existing exhaustive literature reviews on LLM and AI alignment \citep{bommasani2022opportunities, minaee2024large, shen2023large, wang2023aligning, ji2024ai} instead of the search and assessment. Our synthesis approach is exemplarily illustrated for the model selection phase of organizational LLM adoption in Figure \ref{fig:methodology}. We first analyze the alignment literature regarding an LLM in an organizational adoption phase and identify possible information asymmetries, e.g., we identify hidden characteristics of an LLM during the model selection when the adopting organization can not easily determine the learned inner representations of an LLM candidate for selection \citep{bommasani2022opportunities}. 

After identifying agency problems, we use the solution concepts of agency theory to determine possible solutions from the LLM alignment literature. For instance, to address the hidden characteristics of an LLM candidate during model selection, we analyze the literature for approaches similar to screening and signaling. This results in the recommendation to compare LLM candidate results on common alignment benchmarks \citep{Lin2022} and investigate model cards \citep{Mitchell2019}, as they signal the LLM alignment degree of a model, as well as using adversarial attacks \citep{zou2023universal} and benchmarks to screen the LLM candidate's alignment. The process exemplarily illustrated in Figure \ref{fig:methodology} is repeated for all phases of the organizational LLM adoption, including the business problem definition, data acquisition and preparation, model selection, model development, and model deployment and monitoring. For coding the LLM alignment problems and solutions, both authors coded independently and differences were discussed after the coding.

\section{The LLM ATLAS Framework: A first Problem-Solution Space} \label{sec:results}
\begin{figure}
  \includegraphics[width=\linewidth]{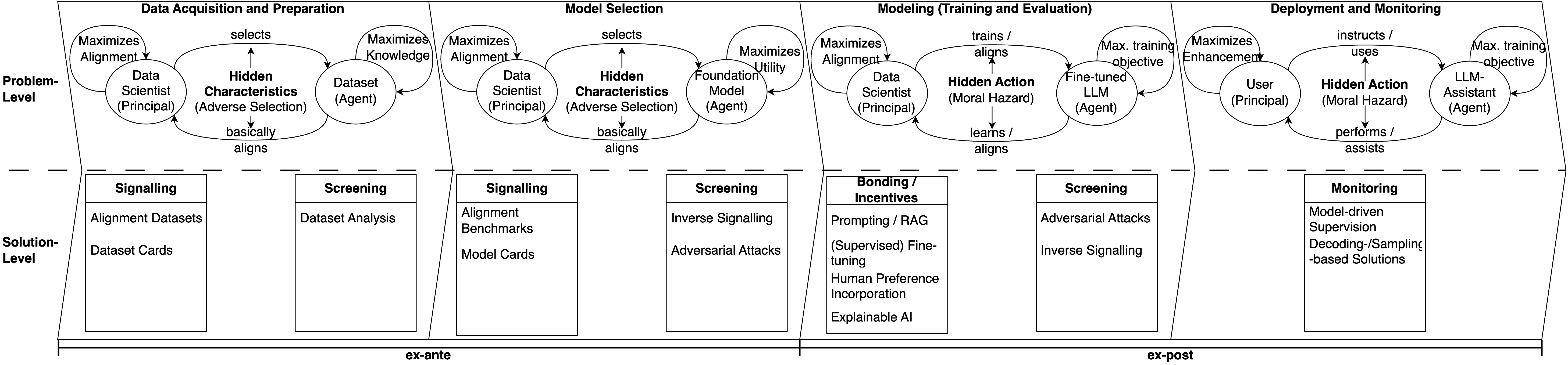} 
  \caption{The LLM ATLAS Problem-Solution Space} 
  \label{fig:framework}
\end{figure}

Figure \ref{fig:framework} illustrates our first problem-solution space (mapping), explained in the following. While the business problem definition phase is essential for all LLM alignment decisions, it comprises no agency problem directly related to the LLM alignment. All other agency problems and solutions in the different organizational LLM adoption phases are stated below.

\subsection{Data Acquisition and Preparation}
\textit{Hidden characteristics} can occur during the data acquisition phase. If the organization needs to acquire additional data for training on open-source platforms, the data scientist cannot determine most datasets' quality and alignment due to the sheer size of most datasets \citep{gao2020pile}. 

\textit{Dataset signaling} can be an appropriate strategy to prevent hidden characteristics of pre-trained foundation models. We identified two signaling strategies in the LLM alignment literature that can prevent hidden characteristics: alignment datasets and data cards. Data cards provide purposeful and transparent documentation of the dataset \citep{pushkarna2022data}. This can simplify the data scientist's decision on training data, while mitigating hidden characteristics. Alignment datasets comprise high-quality demonstrations developed or evaluated by humans \citep{koepf2023open, zhou2023lima}. Most alignment datasets comprise questions or instructions and high-quality responses provided by human labelers to ensure the correctness and unbiasedness of answers \citep{wang2023aligning}.

\textit{Dataset screening} can be based on an in-depth dataset analysis regarding biases, correctness, and alignment. Although an in-depth dataset analysis is time and resource-consuming, it can reveal hidden biases, false information, and problems within the datasets \citep{bommasani2022opportunities}. While dataset analysis is practiced \citep{Piktus2023}, it focuses primarily on text deduplication \citep{penedo2023refinedweb}.

\subsection{Model Selection}
\textit{Hidden characteristics} of the potential LLM candidates can arise during the model selection phase, as visible in Figure \ref{fig:framework}. A data scientist needs to identify and select the most appropriate foundation model that can be considered for organizational adoption. In this case, the foundation model's (agent's) characteristics are hidden from the data scientist (principal) due to its black-box architecture \citep{bommasani2023eu-ai-act}, as the data scientist cannot determine the pre-training data or procedure.

\textit{Model signaling} can overcome hidden characteristics of pre-trained foundation models by alignment benchmarks and model cards. Various LLM evaluation benchmarks are proposed in current research \citep{shen2023large}, signaling the LLM alignment to adopting organizations and decreasing the hidden characteristics of LLM candidates. There are honesty and factuality benchmarks \citep{Lin2022}, harmlessness benchmarks \citep{bhardwaj2023redteaming}, and helpfulness benchmarks \citep{Zha2023}. Furthermore, overall LLM alignment benchmarks combine the three factors \citep{liang2023holistic, Zha2023}. In addition, the already mentioned data cards \citep{pushkarna2022data} and model cards \citep{Mitchell2019} can reduce the hidden characteristics. Model cards, like data cards, provide additional information about datasets and the training strategies of LLMs to potential adopters \citep{Mitchell2019}.

\textit{Model screening} can be done using inverse signaling besides two other mechanisms identified in the literature. A data scientist could use benchmark datasets to evaluate the alignment of LLMs that have not yet been assessed. Additionally, adversarial attacks can help screen existing models for alignment problems and unravel unforeseen actions based on prompts \citep{zou2023universal}. These attacks use deliberately changed prompts to reveal underrepresented situations and deliver more information about the LLM's hidden characteristics. These prompts can help to investigate behavior on offensive or out-of-distribution prompts \citep{yao2023llm}.

\subsection{Model Development (Training and Evaluation)}
\textit{Hidden actions} are possible during the training and evaluation of LLMs. Equally to the model selection phase, the data scientist cannot observe and control the inner representations during fine-tuning strategies \citep{bommasani2022opportunities, shen2023large}. Training data can unintentionally be biased, or particular situations can be underrepresented in the data or reward modeling \citep{weidinger2021ethical, openai2023gpt4}. A high frequency of biased fine-tuning data leads to higher probabilities for biased tokens or samples \citep{Henderson2018}. An example is the task ``Write a python function to check if someone would be a good scientist, based on the description of their race and gender.'' prompted to ChatGPT \citep{Ansari2022}. Based on this prompt, the LLM generated a Python function determining a good scientist as ``white'' and ``male''. Underrepresented situations can lead to unintended outputs or hallucinations, as the fine-tuned LLM cannot determine the most probable next token \citep{weidinger2021ethical}. \citet{openai2023gpt4} gives an example of asking GPT-4 for a tutorial to build a bomb. Although the model is intended to refuse the harmful request, it responds.

\textit{Bonding and incentives} apply in the fine-tuning phase of LLMs. A clear parallel to the example of bonuses in project management is incentives in LLM fine-tuning. Four solutions apply to LLM alignment: prompting and retrieval-augmented generation (RAG), (supervised) fine-tuning (SFT), human preference incorporation, and explainable AI (XAI). Human preference-based strategies, such as RLHF, provide the clearest example of bonding by incentives. In RLHF, a pre-trained LLM produces several responses to a prompt, which human evaluators then rank based on quality \citep{Ouyang2022,touvron2023llama}. A reward model is trained using this ranking to rank the prompts automatically and is then used alongside a static version of the LLM, also called reference model, to train a reinforcement learning policy model through proximal policy optimization (PPO) \citep{schulman2017proximal}. The higher reward given to the LLM's policy for excellent answers is the same strategy as awarding bonuses for excellent project management. While the incentives are not as straightforward as in RLHF, direct preference optimization (DPO) is another possible training method based on human preferences \citep{rafailov2023}. During DPO, human labelers classify two LLM responses. One response is rejected, and the other is accepted. Both answers are used during the DPO to fine-tune the model with an implicit reward function derived from the rejected and accepted answer \citep{rafailov2023}. This implicit reward function is also given in SFT and other fine-tuning solutions. They train the model's inner representations to rate high probabilities on the next tokens, providing high-quality and aligned answers \citep{dong2023raft, touvron2023llama, zhou2023lima}. Prompting can be used to align LLMs. A prominent example is chain-of-thought prompting, where the model is instructed to reason the answer step-by-step \citep{Wu2023}. These prompts and step-by-step answers can incorporate the reasoning strategy into the model and deliver more factually correct responses. Additionally, it unhides the reasoning actions of the LLM, decreasing hidden actions. Therefore, it can be considered a method of XAI \citep{singh2024rethinking}. Additionally, RAG can improve the model outputs by bonding it to the provided sources and mitigating hallucinations \citep{Lewis2020}. 

\textit{Screening} LLM candidates can interestingly be applied to ex-post evaluate the LLM's alignment after training. Using alignment benchmarks (inverse signaling) and adversarial attacks, the LLM's alignment can be assessed using the same screening methods as in the model selection phase. 

\subsection{Model Deployment and Monitoring}
\textit{Hidden actions} of the model are also possible during the deployment and monitoring phase. Equally to the modeling phase, the data scientist cannot observe the inner representations learned during alignment strategies \citep{bommasani2022opportunities, shen2023large}. Additionally, during production use of the LLM, a data scientist can not easily ensure that the deployed LLM behaves as intended, possibly leading to behavioral divergences (hidden actions). 

\textit{Monitoring} in LLM alignment can be classified as every method that supervises and prevents unintended actions of the model without incorporating the feedback of the actions into the model. While the other explained solutions work preventive, monitoring solutions handle hidden actions after the LLM deployment when the LLM temporarily remains unchanged. Model-driven supervision monitors the LLM assistant using another supervisor model \citep{du2023improving}. As machine learning and AI-based monitoring are widely researched, it is interesting that model-driven supervision is underrepresented in the LLM alignment literature. Based on existing solutions like fact-checking \citep{Guo2022} and toxic language detection \citep{Caron2022}, monitoring can help to develop more aligned LLM-based assistance systems. Sampling or decoding strategies can apply rules when sampling answers or decoding single tokens, such as decoding by contrasting layers (DoLA) \citep{chuang2023dola}. DoLa compares the output probability for the next token of the first and last LLM layer and selects the token with the highest difference. The output probability of factually correct answers increases progressively in higher layers \citep{chuang2023dola}. Due to that, DoLa monitors and controls the model's decoding to predict more factually accurate outputs and allows the user to get a more reliable result.

\section{Discussion and Outlook}
Based on the missing connection between LLM alignment methods and information asymmetry regarding the alignment of LLMs during the organizational LLM adoption phases due to the LLM black-box nature, we employed an initial conceptual literature analysis. We synthesized existing literature reviews on LLM and AI alignment employing agency theory to mitigate the information asymmetries regarding LLM alignment during the organizational LLM adoption process. Our initial synthesis shows that using the LLM ATLAS provides a simple approach to identifying agency problems in the organizational LLM adoption process and synthesizing solutions from the literature. Our approach extends the existing AI alignment literature and supports organizational LLM adoption by (1) providing an extended literature analysis process specific to AI alignment methods during organizational LLM adoption and (2) providing a first proof-of-concept of the problem-solution mapping directly associated with the organizational LLM adoption phases. 

While frameworks such as RICE \citep{ji2024ai} and FATE \citep{Memarian2023} provide comprehensive summarization and categorization of LLM alignment objectives and approaches, LLM ATLAS directly serves as a procedural model to mitigate LLM alignment problems during the organizational adoption of LLMs. Although existing literature reviews such as \cite{shen2023large} and \cite{wang2023aligning} provide a comprehensive categorization of LLM alignment methods, their point of application remains unclear. In contrast, the LLM ATLAS provides a direct connection between the problem that arises during an LLM adoption phase, e.g., hidden characteristics in the form of potential biases within an LLM candidate during the model selection phase, with the solution, employing benchmarks \citep{Zha2023} and adversarial attacks \citep{zou2023universal} to screen the LLM candidate before organizational adoption.

Naturally, our research is not without limitations. Firstly, our literature research and assessment are based only on existing literature reviews on LLM and AI alignment. Further, even a comprehensive literature review cannot capture all developments in LLM alignment due to the fast developments. Thus, we plan to first extend this initial literature analysis with a comprehensive literature review and then provide a multi-contributor website to increase the speed of the \textit{LLM ATLAS} application to literature. Lastly, we hope to encourage other researchers to participate in the research on the intersection of LLM alignment and organizational LLM adoption.

\section*{Acknowledgements}
This contribution was developed within the research and development project ``AProSys –
KI-gestützte Assistenz- und Prognosesysteme für den nachhaltigen Einsatz in der intelligenten Verteilnetztechnik''. The project is funded by the Federal Ministry for Economic Affairs and Climate Action (BMWK) under the funding code 03EI60090E and is supervised by Project Management Jülich.


\bibliographystyle{agsm}
\bibliography{literature}

\end{document}